\documentclass[times,twocolumn,final,authoryear]{elsarticle}

\usepackage{ycviu}
\usepackage{framed,multirow}

\usepackage{amssymb}
\usepackage{amsmath}
\usepackage{amsfonts}
\usepackage{mathrsfs}
\usepackage{latexsym}

\usepackage{url}
\usepackage{xcolor}
\definecolor{newcolor}{rgb}{.8,.349,.1}

\journal{Computer Vision and Image Understanding}

\begin{document}

\begin{frontmatter}

\title{Rethink Arbitrary Style Transfer with Transformer and Contrastive Learning}

\author[1]{Zhanjie \snm{Zhang}} 
\author[1]{Jiakai \snm{Sun}}
\author[1]{Guangyuan \snm{Li}}
\author[1]{Lei \snm{Zhao}}
\author[1]{Quanwei \snm{Zhang}}
\author[1]{Zehua \snm{Lan}}
\author[1]{Haolin \snm{Yin}}
\author[1]{Wei \snm{Xing}}
\author[1]{Huaizhong \snm{Lin}}
\author[2]{Zhiwen \snm{Zuo}\corref{cor1}}
\cortext[cor1]{Corresponding author.}
\ead{zzw@zjgsu.edu.cn}
\address[1]{College of Computer Science and Technology, Zhejiang University, No. 38, Zheda Road, Hangzhou 310000, China}
\address[2]{College of Computer Science and Technology, Zhejiang Gongshang University, No. 18 Xuezheng Street, Hangzhou 310018, China}
\received{1 May 2013}
\finalform{10 May 2013}
\accepted{13 May 2013}
\availableonline{15 May 2013}
\communicated{S. Sarkar}

\begin{abstract}
Arbitrary style transfer holds widespread attention in research and boasts numerous practical applications. The existing methods, which either employ cross-attention to incorporate deep style attributes into content attributes or use adaptive normalization to adjust content features, fail to generate high-quality stylized images. In this paper, we introduce an innovative technique to improve the quality of stylized images. Firstly, we propose Style Consistency Instance Normalization (SCIN), a method to refine the alignment between content and style features. In addition, we have developed an Instance-based Contrastive Learning (ICL) approach designed to understand the relationships among various styles, thereby enhancing the quality of the resulting stylized images. Recognizing that VGG networks are more adept at extracting classification features and need to be better suited for capturing style features, we have also introduced the Perception Encoder (PE) to capture style features. Extensive experiments demonstrate that our proposed method generates high-quality stylized images and effectively prevents artifacts compared with the existing state-of-the-art methods.
\end{abstract}


\end{frontmatter}


\begin{figure*}[t]
	\centering
	\includegraphics[width=2.06\columnwidth]{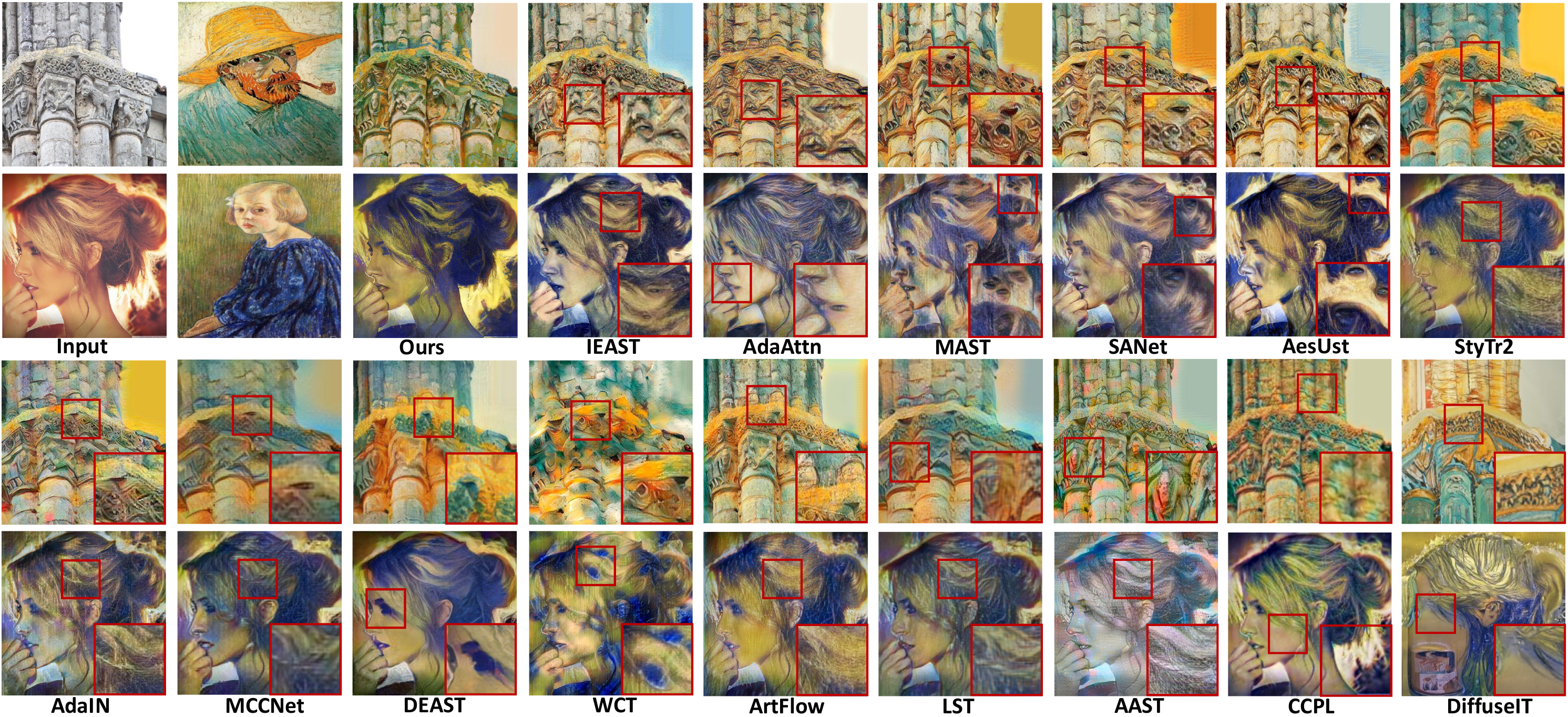} 
	\caption{Stylized examples of the existing arbitrary style transfer method. Although the attention-based arbitrary style transfer method can learn local texture and content-style correlation, they sometimes bring in the content feature of style images in Row 1. Non-attention-based arbitrary style transfer failed to learn detailed texture and also generated artifacts.}
	\label{fig2}
\end{figure*}

\begin{figure*}[htbp]
	\centering
	\includegraphics[width=2.1\columnwidth]{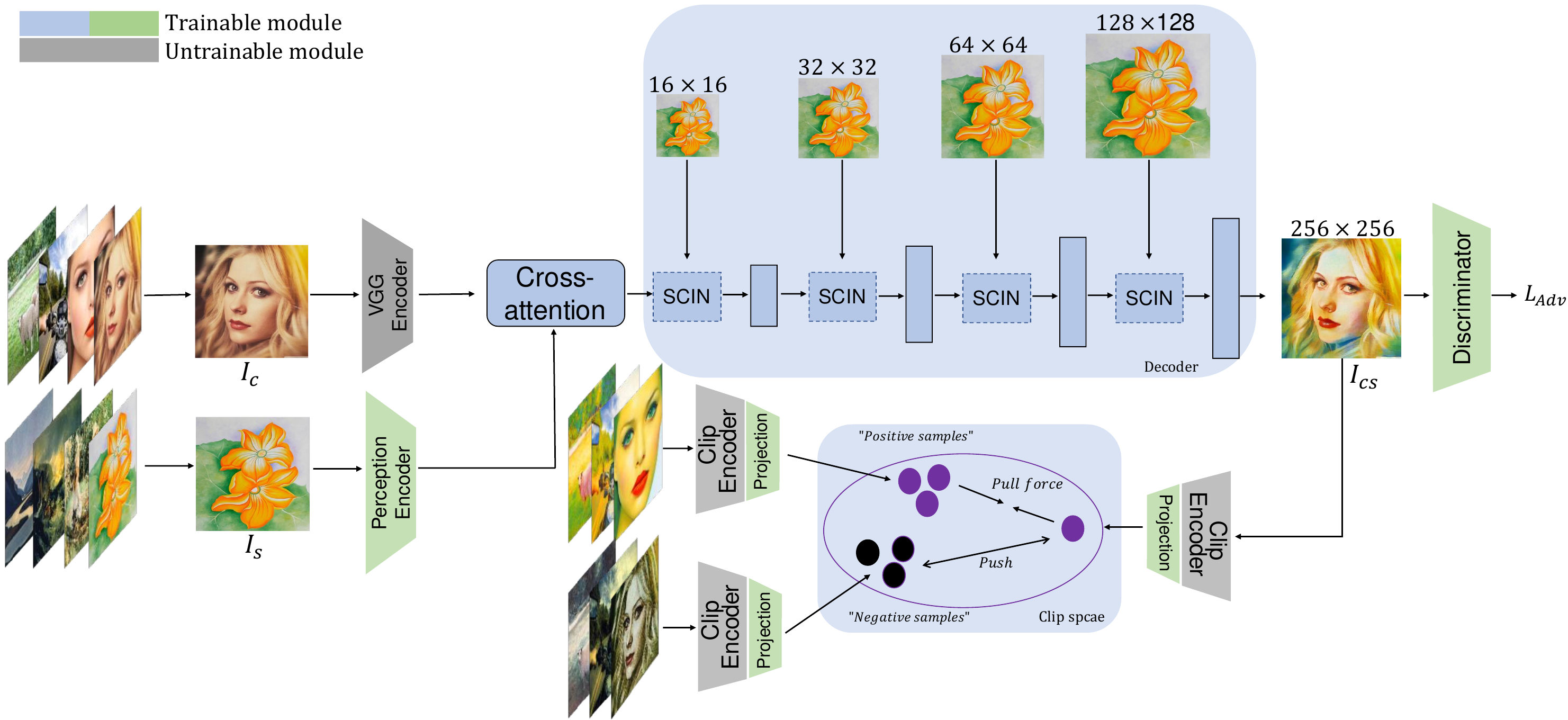} 
	\caption{The overview of the proposed method which consists of pre-trained VGG and Clip encoder, Style Consistency Instance Normalization (SCIN) and discriminator.
	}
	\label{architecture}
\end{figure*}
\section{Introduction}
Style transfer refers to the process of generating a new image that retains the content of a given content image while incorporating the style of a given style image.
Gatys et al.~\citep{gatys2015neural,cai2023image} proposed a seminal work introducing an online optimization method for style transfer. The proposed method used a fixed model (VGG)~\citep{simonyan2014very} to extract content and style features. It iteratively optimized the content features to match the content of the input image and the style features to match the style of the reference image. Additionally, multi-style methods~\citep{dumoulin2016learned,chen2017stylebank} have expanded the capability of style transfer by using a single pre-trained model to perform style transfer with multiple style references.
Then, arbitrary style transfer methods~\citep {huang2017arbitrary} become a research hot spot. Arbitrary style transfer methods can generate stylized images from arbitrary content and style images based on a single model. 

Existing arbitrary style transfer methods can be divided into two categories: (a) attention-based style transfer methods. (b) non-attention-based style transfer methods. 
As representatives of the former, IEAST~\citep{chen2021artistic}, AdaAttN~\citep{liu2021adaattn}, MAST~\citep{deng2020arbitrary}, SANet~\citep{park2019arbitrary}, 3DPS~\citep{mu20223d}, RAST~\citep{ma2023rast} and AesUST~\citep{wang2022aesust} utilize attention mechanisms to locally fuse style features into
content features. While attention-based arbitrary style transfer is highly superior in learning local texture and content-style semantic correlation, these methods often bring evident artifacts, making their results easily
distinguishable from real paintings. Although StyTr2~\citep{deng2022stytr2} use a transformer to avoid introducing artifacts, dropping high-frequency content and style information. As representatives of the latter, AdaIN~\citep{huang2017arbitrary}, DEAST~\citep{zhang2022domain}, WCT~\citep{li2017universal}, ArtFlow~\citep{an2021artflow}, LST~\citep{li2019learning}, Gating~\citep{yang2022gating}, Caster~\citep{zhang2023caster}, UAST~\citep{cheng2023user}, TeSTNerf~\citep{chen2023TeSTNeRF}, AAST~\citep{hu2020aesthetic}, CCPL~\citep{wu2022ccpl} and MCCNet~\citep{deng2021arbitrary} transform the
content features to match the second-order global statistics of style
features without considering local distribution. They may often integrate messy style textures and patterns
into the content target. DiffuseIT~\citep{kwon2022diffusion} utilizes a pre-trained ViT model to guide
the generation process of DDPM models in terms of content preservation and style information changes. Compared to stylized images, artworks can be distinguished as true due to their artistic characteristics, such as colors, strokes, tones, textures, etc. 
The methods above cannot generate high-quality images that possess these artistic elements (See Fig.~\ref{fig2}).

To solve these problems, we propose Style Consistency Instance Normalization (SCIN) to align content features with style features from feature distribution, which helps to supply global style information.
Specifically, we use a transformer~\citep{vaswani2017attention,lyu2023multicontrast,lyu2023region,li2022wavtrans,li2022transformer} as a global style extractor to capture non-local, long-range dependencies of style information from the style image. The transformer outputs scale and bias parameters, which are used to adjust the global information of the content features and match the feature distribution of the style image. Existing style transfer methods often use a content loss and a style loss to ensure the content-to-stylization and style-to-stylization relations, respectively. However, they tend to neglect the stylization-to-stylization relations, which are also crucial for style transfer. Based on this analysis, we propose a novel Instance-based Contrastive Learning (ICL) that can pull the multiple
stylization embeddings closer to each other when they share the same content or style but push far
away otherwise. Besides, we have observed that existing style transfer methods use VGG as the feature extractor, which is trained on the ImageNet Dataset~\citep{deng2009imagenet} and is effective at extracting classification features but not suitable for removing style features (See Fig.~\ref{visual}). Expired by Inception Transformer~\citep{siinception}, we propose a Perception Encoder (PE) to extract style information and avoid paying too much attention to remarkable classification features. 


To summarize, the main contribution of this paper is as follows:
\begin{itemize}
	\item We propose a novel Style Consistency Instance Normalization (SCIN) to capture long-range and non-local style correlation. This can align the content feature with the style feature instead of the mean and variance computed by fixed VGG.
\item Considering existing methods always generate low-quality stylized images with artifacts or stylized images with semantic errors, we introduce a novel Instance-based Contrastive Learning (ICL) to learn stylization-to-stylization relation and remove artifacts. 
\item We analyze the defects of attention-based arbitrary style transfer due to fixed VGG and propose a novel Perception Encoder (PE) that can capture style information and avoid paying too much attention on the remarkable classification feature of style images.
\item Compared to the state-of-the-art method, extensive experiments demonstrate our proposed method can learn detailed texture and global style correlation and remove artifacts.
\end{itemize}

\section{Related Work}
\textbf{Arbitrary Style Transfer}. Recent arbitrary style transfer methods can be divided into two categories: attention-based style transfer methods and non-attention-based style transfer methods. The common idea of the
former category is to apply feature modification globally. Dumoulin et al. propose conditional instance normalization(CIN)~\citep{gulrajani2017improved} to scale and shift the activations in the IN layer. Based on CIN, a simple network can transfer multiple styles. Huang et al.~\citep{huang2017arbitrary} propose to use adaptive affine parameters for arbitrary style transfer. Jing et al.~\citep{jing2020dynamic} propose Dynamic Instance Normalization, but they still depend on VGG's feature space. AdaIN and DIN are very effective methods, but they depend on VGG and can't learn global-style feature maps. 
Park et al.~\citep{park2019semantic} project the style image onto a convolved embedding space to produce modulation parameters. The convolutional neural network itself constrains spade, and convolutional neural networks fail to learn global information and long-range dependence.
Li et al. propose to use feature transforms~\cite {li2017universal}, i.e., whitening and coloring, to directly match content feature statistics to those of a style image in the deep feature space.

For the latter, Park et al. first introduces attention module~\cite {park2019arbitrary} to learn local textures and content-style correlation. AdaAttN introduces a novel AdaAttN module for arbitrary
style transfer. It takes both shallow and deep features
into account for attention score calculation and properly normalizes content features such that feature statistics are well aligned with attention-weighted mean and
variance maps of style features on a per-point basis. Deng et al. propose a transformer-based style transfer framework called
StyTr2
, to generate stylization results with well-preserved
structures and details of the input content image.
InST~\citep{zhang2023inversion} proposes to learn artistic style directly from a single painting and guide diffusion to generate a stylized image. DiffuseIT~\citep{kwon2022diffusion} utilizes a pre-trained ViT model to guide
the generation process of DDPM models in terms of content preservation and style information changes.

\textbf{Contrastive Learning}. Contrastive learning consists of three key ingredients: query, positive examples, and negative samples. The goal of contrastive learning is to make query push negative and pull force positive samples. ContraGAN proposed a conditional contrastive learning loss function to learn sample-to-class and sample-to-sample relations. Chen \emph{et al.}~\citep{chen2021artistic} first build positive and negative samples based on a pre-trained model(VGG) to learn stylization-to-stylization relations. Zhang \emph{et al.}~\citep{zhang2022domain} introduce contrastive learning for style representation using visual features comprehensively to represent style for arbitrary style transfer. Wu \emph{et al.}~\citep{wu2022ccpl} devise a generic contrastive coherence preserving loss to learn local patches. Recently, OpenAI's Contrastive Learning-Image Pre-training (CLIP)~\citep{radford2021learning} built an effective relationship between language and images. CLIPstyler~\citep{kwon2022clipstyler} introduces a text-guided synthesis model that can transfer the style of an image according to specific text.

\textbf{Image-to-Image Translation.} Recently, image-to-image translation~\citep{zhu2017unpaired,lin2020tuigan,chen2021dualast,xu2021drb,zhang2021generating,zhang2024artbank,zhang2024towards,li2024rethinking,zuo2023generative} methods have achieved significant progresses. These methods learn to generate stylized images from the style domain. However, the Per-Style Per-Model method cannot meet the needs of arbitrary style transfer. GAN can learn the statistical information of the style domain, and it has achieved great success in eliminating the artifact of the generated image. Based on this, Lin \emph{et al.}~\citep{lin2021drafting} propose a novel feed-forward style transfer method
named LapStyle. It uses a Drafting Network to transfer global style patterns in low resolution. It adopts
higher resolution Revision Networks to revise local style patterns in a pyramid manner according to outputs of multi-level Laplacian filtering of the content image. Besides, Chen et al.~\citep{chen2021artistic} extend the GAN-based method on arbitrary style transfer. 
DualAST~\citep{chen2021dualast} proposes to use a learnable projection network that maps the VGG feature to learn a dynamic parameter $\alpha$, readjusting affine parameters dynamically. Wang et al. \citep{wang2020learning} introduce white-box cartoon representations to decouple cartoon style transfer into three controllable components and use total-variation loss \citep{aly2005image,sun2023vgos,sun20243dgstream} to impose spatial smoothness on stylized images.
\section{Proposed Method}
The overview of the proposed method is shown in Fig.~\ref{architecture}. We propose to use a pre-trained VGG network to extract content features and a Perception Encoder (PE) to extract style features. Then, we use cross-attention to combine the style features into the content features, enabling the network to learn local style information better. Next, the Spatial and Channel-wise Intelligence Normalization (SCIN) module aligns the content and style features in the spatial domain to ensure that they have global style information. The aligned features are fed into the decoder to reconstruct the stylized image. Additionally, we use a contrastive learning method to learn the relationship between the stylized image and the style image, which complements the content loss and style loss that only considers the relationship between the stylized image and the content image or the style image separately. This allows us to generate high-quality stylized images. 
\begin{figure}[t]
	\centering
	\includegraphics[width=1\columnwidth]{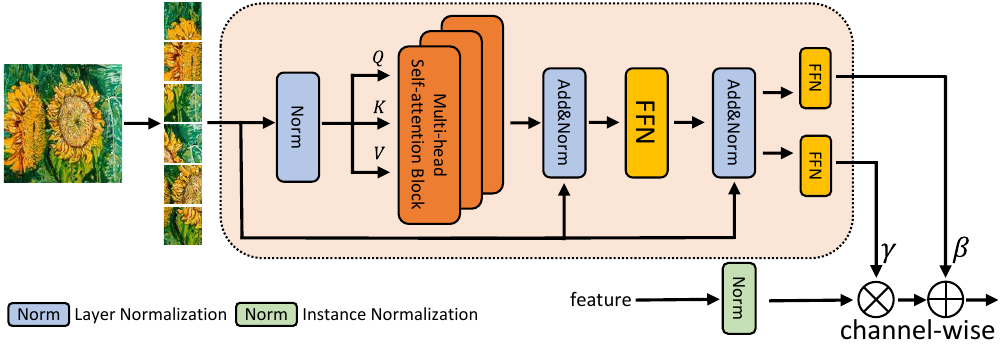} 
	\caption{The structure of our proposed SCIN which mainly consists of
		multi-head self-attention modules (MSA) and a feed-forward
		network (FFN). 
	}
	\label{fig3}
\end{figure}

\begin{figure}[t]
	\centering
	\includegraphics[width=1\columnwidth]{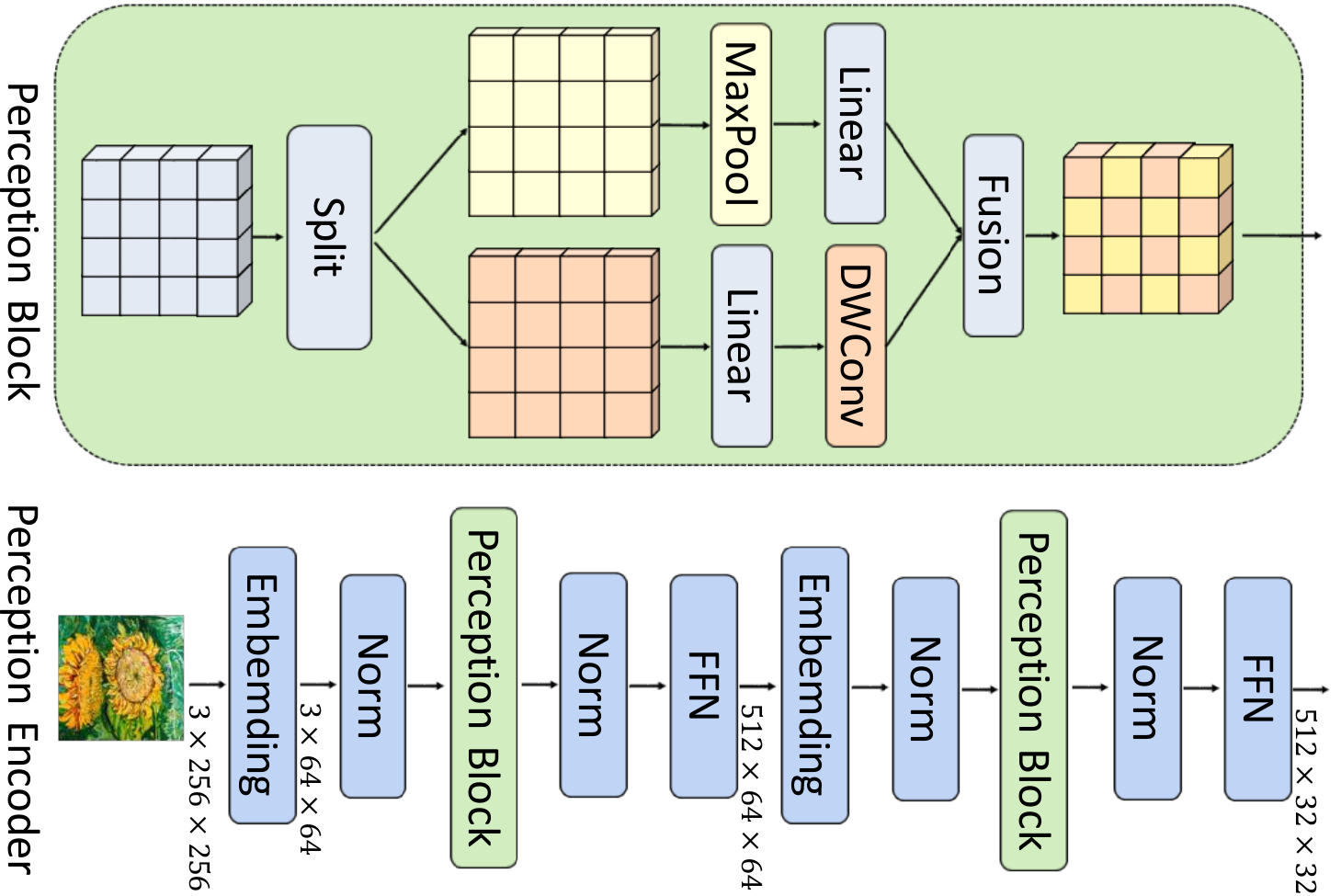} 
	\caption{The detail of Perception Encoder. 
	}
	\label{fig4}
\end{figure}
\subsection{Preliminaries}
\label{3.2}
AdaIN~\citep{huang2017arbitrary} proposes adaptive instance normalization to align the channel-wise mean and variance of content feature $F_{c}$ to match those style feature $F_{s}$. It adaptively computes the affine parameters from the style input:
\begin{equation}
AdaIN(F_{c},F_{s}) =\sigma(F_{s})\left(\frac{F_{c}-\mu(F_{c})}{\sigma(F_{c})}\right)+\mu(F_{s}),
\end{equation}
where  $\mu(x)$ and $\sigma(x)$ are the mean and standard deviation computed across spatial locations.
Given an input batch
$x \in {R}^{N \times C \times H \times W}$, $\mu(x)$ and $\sigma(x)$ can be computed as below:
\begin{equation}
\mu_{n c}(x)=\frac{1}{H W} \sum_{h=1}^H \sum_{w=1}^W x_{n c h w},
\end{equation}
\begin{equation}
\sigma_{n c}(x)=\sqrt{\frac{1}{H W} \sum_{h=1}^H \sum_{w=1}^W\left(x_{n c h w}-\mu_{n c}(x)\right)^2+\epsilon},
\end{equation}

\subsection{Style Consistency Instance Normalization}
Given the embedding of an input style sequence $Z_{s}=(\mathcal{E}_{s 1}, \mathcal{E}_{s 2}, \ldots, \mathcal{E}_{s L})$, we first feed it into the transformer encoder.
The input sequence is encoded into query ($Q$),
key ($K$), and value ($V$):
\begin{equation}
Q=Z_s W_q, \quad K=Z_s W_k, \quad V=Z_s W_v,
\end{equation}
where $W_q$, $W_k$, $W_v$ $\in R^{C\times d_{head}}$, then the multihead attention is calculate by

\begin{equation}
\begin{aligned}
&{F}_{\text {MSA }}(Q, K, V)=\operatorname{Concat}\left(\text { Attention }_1(Q, K, V)\right. \text {, }\\
&\left.\ldots, \text { Attention }_N(Q, K, V)\right) W_o,
\end{aligned}
\end{equation}
where $W_o \in R^{C\times C}$ are learnable parameters, $N$ is the number of attention heads, and $d_{head}=\frac{C}{N}$. Then the encoded style sequency  $Y_{s}$ can be obatained by below:
\begin{equation}
\begin{aligned}
&Y_s^{\prime}={F}_{\mathrm{MSA}}(Q, K, V)+Q, \\
&Y_s={F}_{\mathrm{FFN}}\left(Y_s^{\prime}\right)+Y_s^{\prime},
\end{aligned}
\end{equation}
where ${F}_{\mathrm{FFN}}\left(Y_s^{\prime}\right)=\max \left(0, Y_s^{\prime} W_1+b_1\right) W_2+b_2$ and Layer normalization~\citep{ba2016layer} is applied after each block.
Futhermore, we propose a new SCIN, Let $\gamma^{i}_{s}$ and $\beta^{i}_{s}$ convert $I_{s}$ to the sacling and bias in the $i-th$ activation map.
\begin{equation}
\begin{aligned}
\gamma^{i}_{s}=F^{i}_{FFN}(&Y_s),\beta^{i}_{s}=F^{i}_{FFN}(Y_s),
\end{aligned}
\end{equation}
Given a pair of content image $I_{c}$ and style image $I_{s}$ as input, the proposed SCIN layer can be modeled as:
\begin{equation}
SCIN(F_{c},F_{s}) = \gamma^{s} \times \operatorname{IN}\left(\mathcal{F}_c\right) + \beta^{s},
\end{equation}
where $\gamma^{s}$,$\beta^{s}$ $\in{R}^{N \times C \times 1 \times 1}$. Unlike AdaIN~\citep{huang2017arbitrary}, the scaling $\gamma^{i}_{s}$ and bias $\beta^{i}_{s}$ are learnable variables from global style information. 

In our proposed method, we realign $F_{cs}$ with $\bar{F_{s}}$ based SCIN as below:
\begin{equation}
\begin{aligned}
\bar{F_{cs}} = SCIN(F_{cs},\bar{F_{s}}),\\
\end{aligned}
\end{equation}

In the decoder, every intermediate feature $F_{cs}$ will be aligned with style feature $F_{s}$. We redefine the reconstructed stylized image. $I_{cs}$ as below: 
\begin{equation}
\begin{aligned}
I_{cs} = D(\bar F_{cs},Transformer(I^{1:x}_{s})),\\
\end{aligned}
\end{equation}
where $x=4$, $I^{1:x}_{s}$ represent multi-scale style images in Fig.~\ref{architecture}.

\subsection{Instance-based Contrastive Learning}
\label{3.3}
Contrastive learning~\citep{wu2021contrastive,santa2018visual,chen2020improved,han2021dual} has been used in many fields which can preserve the content~\citep{han2021dual} of the content image and enhance the style~\citep{chen2021artistic} of stylized image. A novel ICL is proposed to learn stylization-to-stylization relations. Instead of previous contrastive learning based on VGG~\citep{chen2021artistic}, we utilize an image encoder of CLIP to obtain instance-based latent code space to improve stylized images. CLIP~\citep{radford2021learning} is an image-text model in which the text encoder and image encoder can project image and text into the same space. So, every image will have its special clip space, including content and style information. So, this is more suitable for learning stylization-to-stylization relations than VGG. Assume the $b^{n}_{s}$ and $b^{n}_{c}$ represent the bach size number $n$ of style images and content images. For every style image $s_{j}$ and content image $c_{i}$ of $b^{n}_{s}$ and $b^{n}_{c}$, i.e. $i\in[0,n-1], j\in[0,n-1]$. We use $s_{i}c_{j}$ to denote the corresponding stylized images. For all the content images and style images, we build stylized images as ${s_{1}c_{1},s_{1}c_{2},...,s_{1}c_{n};...; s_{n}c_{1},s_{n}c_{2},...,s_{n}c_{n}}$. For every stylized image $s_{i}c_{j}$, we build ``Positive Examples:" $s_{m}c_{n}$, i.e. $m=i, n\neq j$ and ``Negative Samples": $s_{m}c_{n}$, i.e. $m\neq i, n\neq j$. Based on these, Instance-based Contrastive Learning can be calculated by:
\begin{center}
	\begin{equation}
	\begin{aligned}
	L_{contra}= L_{pos} + L_{neg},\qquad\qquad\\
	L_{pos}= -log(\frac{P_s}{P_s+N_s}),L_{neg}= -log(\frac{P_c}{P_c+N_c}),\\
	P_s = exp((M_s(s_{i}c_{j})^{\top}M_s(s_{i}c_{j})/\tau), \\
	N_s =\sum exp(M_s(s_{i}c_{j})^{\top})M_s(s_{i}c_{j})/\tau),\\
	P_c =exp(M_c(s_{i}c_{j})^{\top}M_c(s_{i}c_{j})/\tau),\\
	N_c = \sum exp(M_c(s_{i}c_{j})^{\top}M_c(s_{i}c_{j})/\tau),
	\end{aligned}
	\end{equation}
\end{center}
where $M_s = l_s(E_{clip}(\cdot))$, $M_c = l_c(E_{clip}(\cdot))$, in which $E_{clip}$ represents the image encoder of CLIP, $l_c$ and $l_s$ are the projection network to obtain content code and style code. $\tau$ is a high-parameters to control push and pull force set to 0.3. Based on the above analysis, we propose Instance-based Contrastive Learning to constrain pixel-level according to image latent code space.
\begin{figure*}[htb]	
	\centering
	\includegraphics[width=1\textwidth,height=0.7\textwidth]{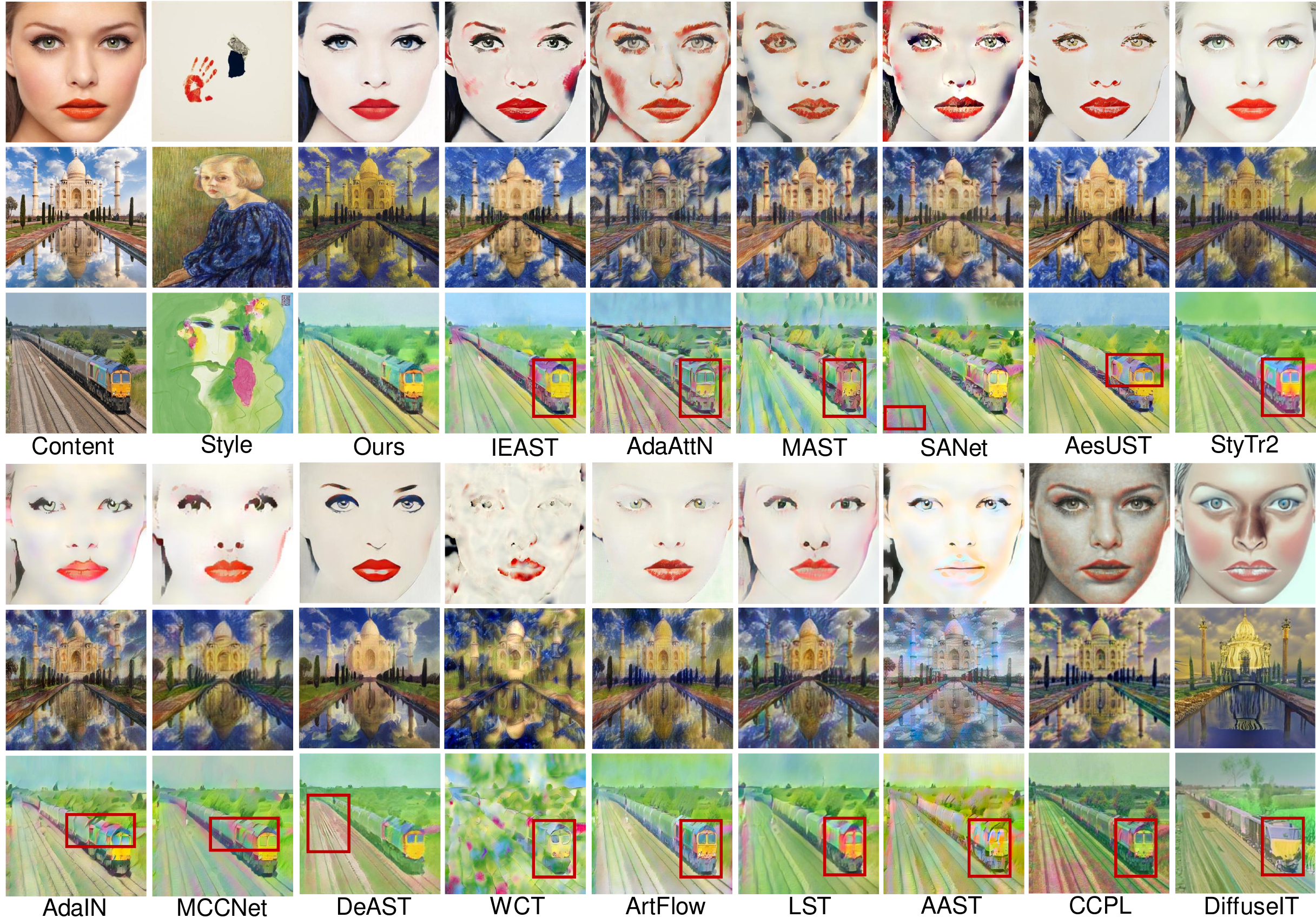} 
	\caption{Qualitative comparison with other state-of-the-art arbitrary style transfer methods.}
	\setlength{\belowcaptionskip}{-200cm}  
	\label{fig5}
\end{figure*}

\begin{figure*}
	\includegraphics[width=\textwidth]{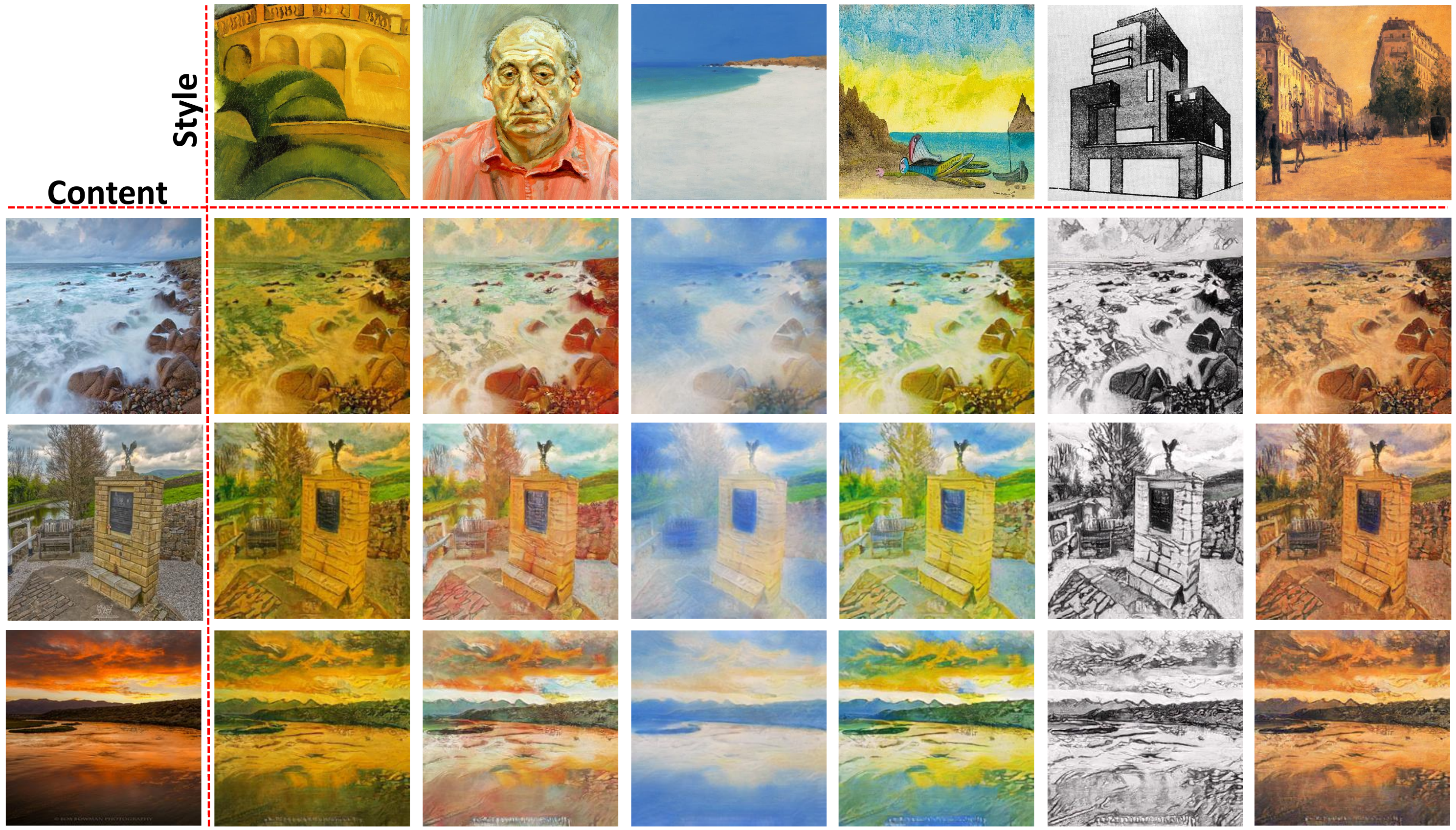}
	\caption{Stylized examples generated by our proposed method. The first row displays several artworks taken from WikiArt, which were used as the style image. The first column shows the content image, while the remaining images are the stylized images produced by our proposed model.}
	\label{tissue}
\end{figure*}
\subsection{Perception Encoder}
\label{3.1}
Transformer~\citep{liu2021swin,cao2021swin,yu2022metaformer,plizzari2021skeleton,sortino2023transformer,wang2022frame} have demonstrated its superiority, especially in long-range dependence and obtaining global information. However, the transformer is weak in high-frequency information processing. We propose Perception Encoder to solve this problem absolutely, and we have verified its superiority in Fig.~\ref{visual}, and detailed architecture is shown in Fig.~\ref{fig4}. Given style image $I_{s}^{3\times 256 \times 256}$, it first embedding into patches $F_{s}^{512\times 64 \times 64}$. The and patches are dovided into $F_{h}^{256\times 64 \times 64}$ and $F_{l}^{256\times 64 \times 64}$ along the channel dimension. We propose a parallel structure to learn the high-frequency components $F_{h1}^{128\times 64 \times 64}$ and $F_{h2}^{128\times 64 \times 64}$. $F_{h1}$ is embedded with
a max-pooling and a linear layer~\citep{szegedy2015going} and $F_{h2}$ is fed into a linear and a depthwise convolution layer~\citep{chollet2017xception,mamalet2012simplifying,sandler2018mobilenetv2}. To calculate the attention map, we divide the input to style feature  $F_{s}^{3\times 16 \times 16}$. We define this process as below:
\begin{equation}
\begin{aligned}
& \boldsymbol{F}_{h 1}=\operatorname{FC}\left(\operatorname{MaxPool}\left(\boldsymbol{F}_{h 1}\right)\right) \\
& \boldsymbol{F}_{h 2}=\operatorname{DwConv}\left(\operatorname{FC}\left(\boldsymbol{F}_{h 2}\right)\right),
\end{aligned}
\end{equation}
where FC represents the fully connected layer. $Y_{h1}$ and $Y_{h2}$ denote the outputs of high-frequency mixers.
For low-frequency mixer,
\begin{equation}
\boldsymbol{F}_l=\operatorname{Upsample}\left(\mathcal{F}_{\mathrm{MSA}}\left(\operatorname{AvePooling}\left(\boldsymbol{F}_l\right)\right)\right) 
\end{equation}
then
\begin{equation}
\boldsymbol{F}_{\boldsymbol{s}}=\operatorname{Concat}\left(\boldsymbol{Y}_l, \boldsymbol{Y}_{h 1}, \boldsymbol{Y}_{h 2}\right)
\end{equation}
Then we repeat the above process by projecting style images into style feature $F_{s}^{512\times 32 \times 32}$.

\subsection{Other loss functions}
\quad\textbf{Perceptual Content and Style Loss}. Following previous style transfer method~\citep{liu2021adaattn,deng2020arbitrary,chen2021artistic,park2019arbitrary}. For the layer $x$ of the VGG encoder, e.g., $E^{x}_{VGG}$, the perceptual content and style loss can be calculated as below:
\begin{equation}
\begin{aligned}
L_{c}= \sum^{L}_{i=4}||E^{x}_{VGG}(I_{cs})
-E^{x}_{VGG}(I_c)||_2,\\
L_{s}= \sum^{L}_{i=1}||\mu (E^{x}_{VGG}(I_{cs})
-\mu(E^{x}_{VGG}(I_s))||_2 \\
+||\theta((E^{x}_{VGG}(I_{cs}))
-\theta(E^{x}_{VGG}(I_s)))||_2,
\end{aligned}
\end{equation}
Where $\mu$, and $\theta$ denotes the channel-wise mean and standard deviation. For $E^{x}_{VGG}$, we use the $ReLU4\_1$ and  $ReLU5\_1$ to compute $L_c$. Differently, we use $ReLU1\_1$, $ReLU2\_1$,$ReLU3\_1$, $ReLU4\_1$ and $ReLU5\_1$ to compute style loss, paying more attention to different scale style feature information.

\textbf{Adversarial Loss}. Inspired by the Generative Adversarial Network (GAN)~\citep{zhu2017unpaired,chen2021artistic}, which can effectively make the data distribution of the stylized image $I_{cs}$ more close to style images $I_{s}$. The adversarial loss can make the stylized images $I_{cs}$ look more realistic and remove non-human visual artifacts. Then, we define Adversarial Loss can be computed as: 
\begin{equation}
L_{Adv}=\underset{y \sim I_s}{E}\left[\log \left(D_s(y)\right)\right]+\underset{x \sim I_{c s}}{E}\left[\log \left(1-D_s(x)\right]\right..
\end{equation}

\textbf{Identity Loss}. Attention-based style transfer tends to lose content structure, and identity loss~\citep{park2019arbitrary,lin2020tuigan,zhao2020unpaired} is used to solve this problem. Following prior identity loss, $L_{identity}$ can be calculated as below:
\begin{equation}
\begin{aligned}
L_{Identity}=\lambda_{identity1}(||I_{cs}-I_c||_2 +||I_{cs}-I_{s}||_2), \\
+ \sum_{i=0}^L\lambda_{identity1}(||E^{x}_{VGG}(I_{cs})-E^{x}_{VGG}(I_c)||_2\\
+||E^{x}_{VGG}(I_{cs})-E^{x}_{VGG}(I_{s})||_2) ,
\end{aligned}
\end{equation}
where $\lambda_{1}=50$, $\lambda_{2}=1$, $ReLU1\_1$, $ReLU2\_1$,$ReLU3\_1$, $ReLU4\_1$ and $ReLU5\_1$ are used.

\subsection{Objective Loss Function}
We summarize all the above losses to obtain the final objective loss function.
\begin{equation}
\begin{aligned}
L=\lambda_{1}L_s +\lambda_{2}L_c+\lambda_{3}L_{identity}+
\lambda_{4}L_{Adv} + \lambda_{5}L_{contra}
\end{aligned}
\end{equation}
where $\lambda_{1}=1$, $\lambda_{2}=1$, $\lambda_{3}=5$, $\lambda_{4}=1$, $\lambda_{5}=0.3$.

\begin{table}[htb]
	\caption{Quantitative comparison with other state-of-the-art arbitrary style transfer methods. ``Times'' means the inference time with a scale of $512\times512$ pixels using single RTX 2080.}
	\centering
	\setlength{\tabcolsep}{0.1cm}
	\begin{center}		
		\begin{tabular}{c|c|cccc}
			\hline
			\footnotesize  &CF& GE+LP& Deception  & Preference &Times
			\\
			\hline
			\footnotesize WikiArt &-&- &0.784&-&-
			\\
			\footnotesize Ours &\bf 0.432&\bf1.615&\bf 0.573 &-&0.153
			\\
			\footnotesize IEAST &0.412&1.405&0.476&0.381&0.064
			\\
			\footnotesize AdaAttN &0.380&1.430&0.372&0.363& 0.142
			\\			
			\footnotesize MAST &0.360& 1.511& 0.324&0.268&0.124 
			\\ 	
			\footnotesize SANet &0.366&1.552 & 0.346&0.312&0.064 
			\\ 	
			\footnotesize AdaIN &0.383&1.552& 0.346&0.227&0.045
			\\ 	
			\footnotesize CAST &0.335& 1.581&0.423&0.376&0.045
			\\ 	
			\footnotesize WCT &0.346&1.540  & 0.172&0.152&0.408 
			\\
			\footnotesize LST &0.408& 1.459 & 0.408&0.350& \bf 0.036 
			\\
			\footnotesize StyTR2 &0.336&1.602&0.545 &0.392& 2.781
			\\
			\footnotesize AesUst &0.420&1.524  &0.568 &0.415& 0.066
			\\
			\footnotesize ArtFlow &0.411&1.505  &0.356&0.350&0.382
			\\
			\footnotesize AAST &0.368&1.461  &0.307 &0.263& 1.153
			\\
			\footnotesize CCPL &0.422&1.594  &0.427 &0.393& 0.043
			\\
			\footnotesize MCCNet &0.382&1.473  &0.357 &0.273&0.023
			\\
			\footnotesize DiffuseIT &0.295&1.482  &0.287 &0.252&35.52
			\\
			\hline 
		\end{tabular}
	\end{center}
	\label{table1}
\end{table}

\begin{figure*}[t]
	\centering
	\includegraphics[width=1\textwidth,height=0.32\textwidth]{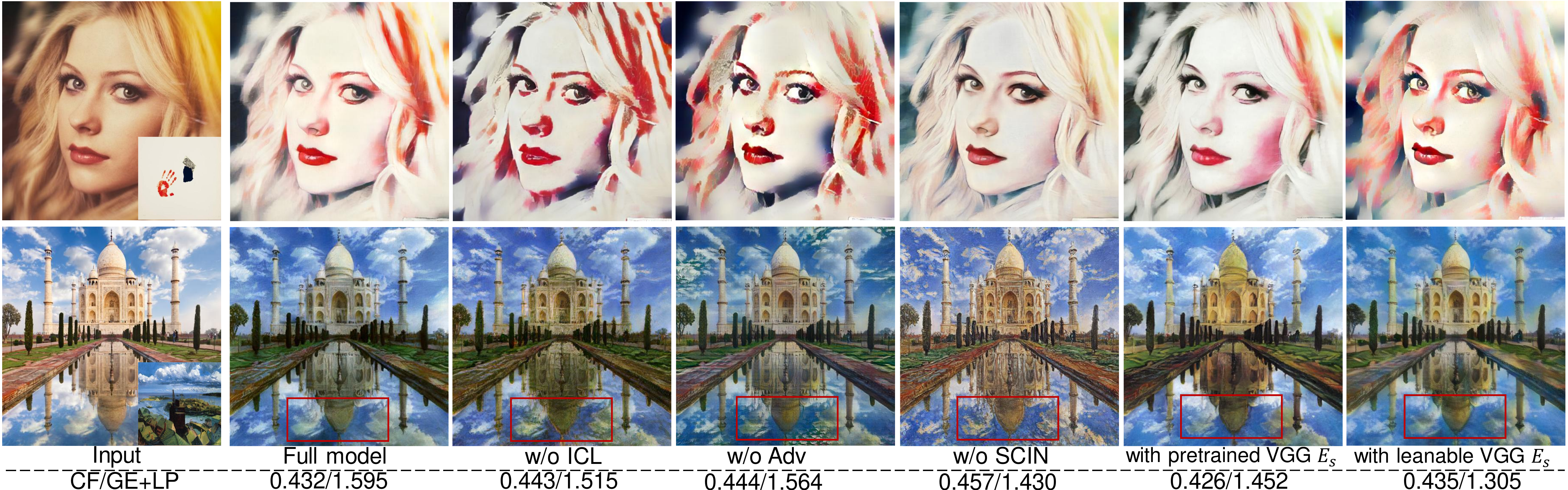} 
	\caption{Ablation study on the different loss functions. Zoom in for better observation. ``with pretrained VGG $E_{s}$'' represents using fixed VGG as style encoder and ``with learnable VGG $E_{s}$'' denotes learnable VGG as style encoder.}
	\label{fig7}
\end{figure*}

\section{Experiments}
\subsection{Implementation Details}
The content and style images come from MS-COCO~\citep{lin2014microsoft} and WikiArt~\citep{wikiart}. During training, all 82,783 content images and 79,433 style images are resized to $512\times512$ pixels. Before feeding into a generator, they are cropped to $256\times256$ pixels. For inference, some style pictures, including oil paintings, watercolor paintings, sketches, etc., are randomly collected. In addition, we also collect some pictures, including buildings, forests, and live photos, as content images and keep their original size sent to our proposed model to generate stylized images. 
\subsection{Comparisons with SOTA Methods}
As shown in Fig.~\ref{fig5}, we compare the proposed method with the state-of-art method, including attention-based style transfer: IEAST~\citep{chen2021artistic}, AdaAttn~\citep{liu2021adaattn}, MAST~\citep{deng2020arbitrary}, SANet~\citep{park2019arbitrary}, AesUst~\citep{wang2022aesust}, StyTr2~\citep{deng2022stytr2}, non-attention-based style transfer: AdaIN~\citep{huang2017arbitrary}, DEAST~\citep{zhang2022domain}, WCT~\citep{li2017universal}, ArtFlow~\citep{an2021artflow}, LST~\citep{li2019learning}, AAST~\citep{hu2020aesthetic}, MCCNet~\citep{deng2021arbitrary}, CCPL~\citep{wu2022ccpl}, DiffuseIT~\citep{kwon2022diffusion}. IEST, AdaAttN, MAST, SANet, and AesUST sometimes introduce undesired semantic structures from the style image to the stylization result. WCT has severe problems with content preservation. AdaIN and MCCNet suffer from content structure blur and style pattern distortion issues. StyTr2, DeAST, ArtFlow, and LST fail to learn local texture from the style image. Although CCPL can effectively maintain the content structure, it cannot learn the content-style semantic correlation. For AAST, there is an obvious style deviation between the style image and the stylized image. DiffuseIT has the problem of content structure consistency and style oversaturation. Compared with these methods, our proposed method can generate stylized images with content structure and style texture and does not introduce unexpected semantic structures. Besides, we randomly chose high-quality stylized images, as shown in Fig.~\ref{tissue}.

\subsection{Qualitative Comparisons}
\textbf{CF, GE, and LP Scores.} Wang \emph{et al.}~\citep{wang2021evaluate} proposed three novel evaluation metrics to evaluate the quality of style transfer: content fidelity (CF), global effects (GE), and local patterns (LP). Specifically, CF is used to measure the faithfulness of content characteristics and 
GE assesses the stylization of global colors and textures. LP assesses the stylization quality in terms of the similarity of the local style. Higher factors mean a better style transfer result. This section uses 5 content and 10 style images for other state-of-the-art methods to compare with our proposed method. The visually compared samples and metrics are shown in Fig.~\ref{fig5} and Tab.~\ref{table1}.

\textbf{Deception Score.} Deception scores represent whether a user can distinguish stylized and authentic art images. Higher scores mean a higher percentage of stylized images, distinguished as real art images. We randomly selected 20 synthesized images for each method and asked 50 subjects to guess.
In order to compare the advantages of our method more intuitively, we randomly select the same number of WikiArt images to ask 50 subjects. As shown in Tab.~\ref{table1}, our proposed method gets higher deception scores, verifying that our proposed method can generate more real stylized images.

\textbf{Preference Score.} Preference Score means popularity between our and other SOTA methods. We conduct A/B Test user studies to compare
the stylization effects of our method with the SOTA method.
We selected randomly 10 content and 15 style images to synthesize 150 stylized images
for each method. Then 20 content-style pairs are randomly selected for each participant, and we show them the stylized images generated by our and the other 15 state-of-art methods. Next, we ask each participant to choose his/her favorite stylization result for each content-style pair.
We finally collected 1600 votes from 80 participants and presented the percentage of votes for each method
in the fifth row of Tab.~\ref{table1}. The results show that users prefer the stylized images generated by our method more.

\begin{figure}[t]
	\centering
	\includegraphics[width=0.95\columnwidth]{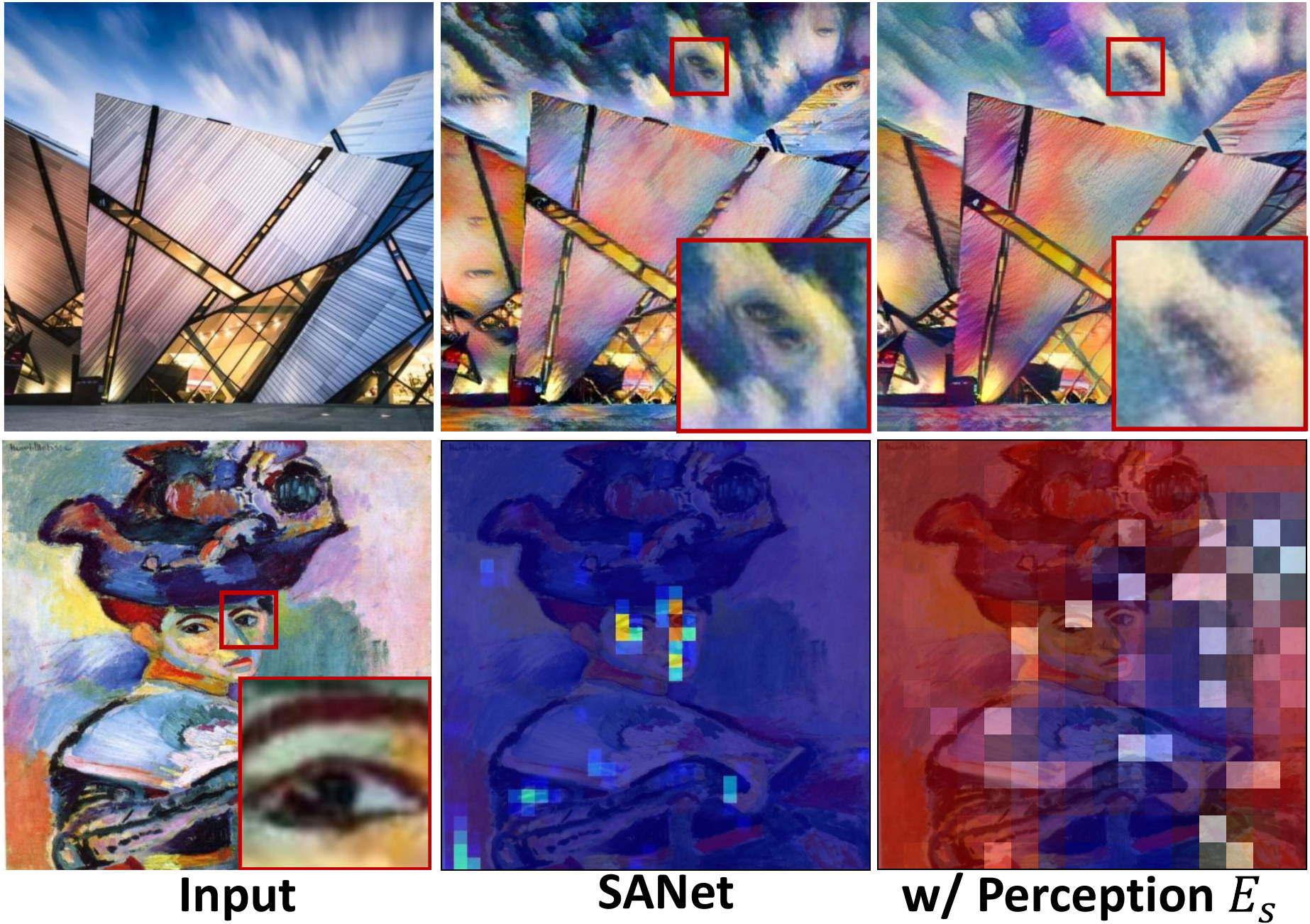} 
	\caption{The reason for the introduction of artifacts in existing style transfer methods..}
	\label{visual}
\end{figure}

\subsection{Abalation Studies}
In this section, we conducted an ablation analysis to assess the effectiveness of the proposed method, as depicted in Fig.~\ref{fig7}.
When $L_{Adv}$ is removed, the generated stylized images show disharmonious patterns and obvious artifacts such as repetitive textures, leading to a significant degradation in quality. This indicates that adversarial training is crucial for generating more harmonious and realistic stylized images. When $L_{ICL}$ is removed, the quality of stylized images degrades in terms of content structure, texture, and brush stroke.The ablation study shows that the SCIN module plays a crucial role in enhancing the style consistency between the stylized images and the style images. Without the SCIN module, the stylized images showed significant style inconsistency. If VGG is used as the style extractor, artifacts may still appear due to its training to extract classification features. Using a trainable VGG structure to extract style information may not be sufficient, resulting in lower quality stylized images.

The impact of the Perception Encoder (PE) is shown in Fig.~\ref{visual}. Existing attention-based methods~\citep{li2023dudoinet,li2023self,li2023rethinking,cui2022attention} usually use SANet as the backbone, which can learn detailed style texture. However, the attention map is calculated from VGG's feature space (i.e., Relu4\_1 and Relu5\_1). VGG can effectively capture remarkable classification features (e.g., eyes). This effectively keeps the content structure of stylized images but does not encode style information. To demonstrate this, we use SANet as a baseline and find that the question comes from the style feature map Relu5\_1 in VGG. So, we visualize the feature map ReLu5\_1. ``w/ Perception $E_{s}$'' using our proposed Perception Encoder instead of fixed VGG. In this part of the experiment, we set the style and content weight to 1 as a baseline. The architecture of the Perception Encoder is shown in Fig.~\ref{fig4}

\section{Conclusion}
This paper proposes a unified network architecture that can generate high-quality stylized images. Specifically, we introduce a novel Style Consistency Instance Normalization (SCIN) method to align the content feature with the style feature in the feature space. Then, we propose an Instance-based Contrastive Learning (ICL) method to learn the stylization-to-stylization relations to improve stylized quality. Additionally, we analyze the limitations of using fixed VGG as a feature extractor and propose a Perception Encoder (PE) to capture style information more effectively. In the future, we plan to explore more general methods to improve the quality of style transfer further.

\section{CRediT authorship contribution statement}
\textbf{Zhanjie Zhang:} Conceptualization, Methodology, Software, Writing – original draft. 
\textbf{Jiakai Sun:} Conceptualization, Methodology, Writing – original draft.
\textbf{Guangyuan Li:} Conceptualization, Methodology, Writing – original draft.
\textbf{Lei Zhao:} Conceptualization, Methodology, Writing – review editing. 
\textbf{Quanwei Zhang}: Software, Investigation, Data curation, Validation, Writing – review editing. 
\textbf{Zehua Lan:}Software, Validation, Data curation, Writing – review editing.
\textbf{Haolin Yin:} Software, Validation, Visualization, Writing – review editing. 
\textbf{Wei Xing}: Methodology, Supervision, Writing – review editing. 
\textbf{Huaizhong Lin}: Methodology, Supervision, Writing – review editing. 
\textbf{Zhiwen Zuo:} Supervision, Writing – review editing.

\section{Declaration of Competing Interest}
The authors declare that they have no known competing financial interests or personal relationships that could have appeared to influence the work reported in this paper.

\section{Acknowledgments}
This work was supported by Zhejiang Elite Program project (2022C01222),  National Natural Science Foundation of China (62172365), the Key Program of the National Social Science Foundation of China (19ZDA197), the Natural Science Foundation of Zhejiang Province (LY21F020005, 2021009, 2019011), MOE Frontier Science Center for Brain Science \& Brain-Machine Integration (Zhejiang University).Brain-Machine Integration (Zhejiang University).

\bibliographystyle{model2-names}
\bibliography{refs}
\end{document}